\documentclass{ifacconf}

\usepackage{amsmath,amssymb,amsfonts}
\usepackage{algorithmic}
\usepackage{graphicx}
\usepackage{textcomp}
\usepackage[dvipsnames]{xcolor}
\usepackage{natbib}
\usepackage{todonotes}
\usepackage{booktabs} 
\usepackage{bm}
\usepackage{siunitx}  
\usepackage[normalem]{ulem} 

\newcommand{\norm}[1]{\left\lVert#1\right\rVert}

\begin{document}
\begin{frontmatter}

\title{
Data-Driven Dynamic Modeling of a Tendon-Actuated Continuum Robot
\thanksref{footnoteinfo}} 

\thanks[footnoteinfo]{This project has received funding from the European Research Council (ERC) under the European Union’s Horizon 2020 research and innovation programme, through the ERC Advanced Grant 101017697-CRÈME.\\
\textsuperscript{\textdagger} H.M. Hansen and B.K. Sæbø contributed equally to this work and should be considered co-first authors.}

\author[First]{Harald Minde Hansen$^\dagger$} 
\author[Second]{Bjørn Kåre Sæbø$^\dagger$} 
\author[Second]{Kristin Y. Pettersen}
\author[Second]{Jan Tommy Gravdahl}
\author[First]{Mario Di Castro}

\address[First]{European Organization for Nuclear Research (CERN), Switzerland \{harald.minde.hansen,mario.di.castro\}@cern.ch}
\address[Second]{Department of Engineering Cybernetics, Norwegian University of Science and Technology, NTNU, NO-7491 Trondheim, Norway \{bjorn.k.sabo, kristin.y.pettersen, jan.tommy.gravdahl\}@ntnu.no}

\begin{abstract}                
Developing dynamic models for tendon-driven continuum robots is challenging due to their nonlinear, high-dimensional, and friction-dominated dynamics. This paper presents a comparative study of data-driven system identification methods—including N4SID, ARX, and SINDYc—for modeling a tendon-actuated continuum robot with rolling joints developed at CERN. Despite the high number of joints of the robot, experimental analysis reveals that a two-degree-of-freedom dynamic model can accurately capture the system dynamics, owing to strong kinematic dependencies between the joints. The models are validated against experimental data, and used in the design of a model predictive controller, demonstrating their feasibility for real-time control.

\end{abstract}


\end{frontmatter}

\section{Introduction}

\begin{figure}
   \begin{center}
   \includegraphics[width=0.9\linewidth]{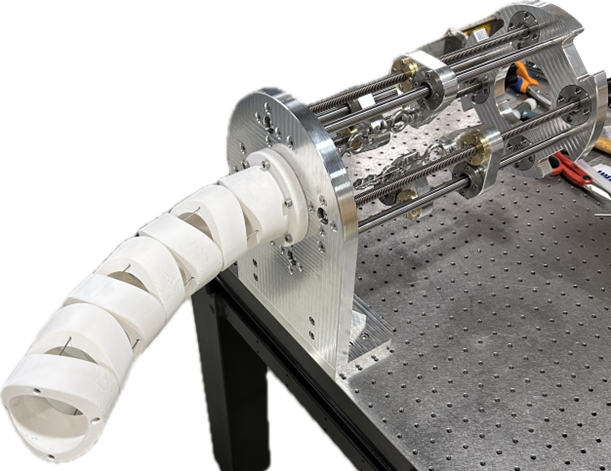}    
   \caption{Tendon-actuated snake robot with rolling joints.}
   \label{fig:snake}
   \end{center}
\end{figure}



The European Organization for Nuclear Research (CERN)  operates underground accelerator facilities where high-energy particle collisions create strong magnetic fields and radioactive environments, limiting human access. To address this, radiation-tolerant robots are deployed for remote maintenance, replacing over 120 manual interventions annually \citep{dicastro2018}. To improve dexterity in confined and complex geometries, a tendon-actuated snake robot is under development. This manipulator consists of rolling, orthogonally arranged links actuated by four continuous cables, enabling flexible shape adaptation through cable tension without mechanical joint coupling.



Mathematical modeling of continuum robots is in general a very challenging task due to their highly nonlinear kinematics and dynamics. Although dynamic models exist, such as \citep{rucker2011a}, they are often computationally expensive and may not be suitable for real-time control. Many works instead focus on static and kinetic models \citep{rao2021} which allow control of the robots in a quasi-static fashion. A Newton-Euler dynamic model of the CERN snake robot was developed in \citep{dantuono2023, dantuono2024}. This resulted in an $N$-Degree of Freedom (DoF) dynamic model, where each DoF represents a joint angle. However, as joints in the same plane are kinematically dependent on each other during normal operations, the $N$-DoF model is redundant under standard operating conditions. It is therefore not possible to control each joint torque to arbitrary references when the tendon tensions are actuated. To obtain a model suitable for real-time dynamic control of the robot, our goal is therefore to utilize data-driven system identification algorithms to develop an efficient lower-dimensional dynamic model of the CERN snake robot, to provide an actuator-to-task mapping.

System identification provides a systematic framework for deriving mathematical models of dynamical systems directly from data, which is particularly valuable for continuum robots where first-principles modeling is often intractable. Classical linear approaches include parametric structures such as ARX models \citep{ljung1999} and subspace identification methods, notably N4SID \citep{vanoverschee1994}, which enable efficient estimation of state-space representations from input–output data. While these methods are well established for linear time-invariant systems, recent advances have extended data-driven identification to nonlinear regimes. In particular, the sparse identification of nonlinear dynamics (SINDY) framework \citep{brunton2016, brunton2016a} enables the discovery of parsimonious governing equations from data, offering interpretability alongside predictive capability. Such methods provide a bridge between purely data-driven and physics-based modeling, especially when combined with structured representations or prior knowledge.

In the context of continuum and soft robotics, where nonlinearities, hysteresis, and high-dimensional deformation complicate analytical modeling, data-driven identification has become increasingly prominent \citep{xu2017, muller2022, bruder2019}. Early efforts demonstrated improved kinematic and dynamic modeling accuracy over classical constant-curvature assumptions, while more recent work has focused on integrating learning with control. The data-driven identification approach of \citep{parvaresh2022} highlights the effectiveness of direct actuator-to-task mappings for continuum arms. Building on this, \citep{zhang2024} propose a real-time hybrid model-based and data-driven nonlinear controller that improves tracking accuracy and robustness. A broader perspective is provided by \citep{liu2025}, who survey data-driven methods for sensing, modeling, and control of soft continuum robots, emphasizing the growing importance of hybrid approaches that combine physical insight with learning. Collectively, these works underline a clear trend toward integrating system identification techniques within adaptive and learning-based control frameworks for continuum robotics.


The main contributions of this paper are as follows: Several linear and nonlinear data-driven modeling approaches are implemented and compared in order to identify an accurate and efficient dynamic model of the CERN snake robot, and a reduced order model is shown to be sufficient to capture the system dynamics. The models are validated against experimental data, and the best-performing models are shown to accurately capture the system behavior. These identified models are used to develop an MPC to evaluate their feasibility for real-time model-based control.


The paper is organized as follows; Sec.~\ref{sec:system} gives an overview of the manipulator system to be modeled, Sec.~\ref{sec:datadriven} briefly summarizes the methods applied for modeling the system, before details on their implementation and validation are presented in Sec.~\ref{sec:experiments}, together with experimental results of MPC implementations on two of the models. Sec.~\ref{sec:conclusions} concludes the paper and discusses future work.

\section{ROBOT SYSTEM OVERVIEW}
\label{sec:system}
This section presents the recently developed tendon-actuated continuum robot at CERN, giving an overview of its kinematic and dynamic properties.

The CERN snake robot is a tendon-actuated robot with rolling joints, see Fig.~\ref{fig:snake}. It consists of $N+1$ links, whose configuration is described by the vector of joint angles $\bm{q} \in \mathbb{R}^N$, with $N=6$ joints for the current prototype. The robot is actuated by four tendons, each connected to a screw-drive motor mounted at the system base.  These tendons traverse through all the links to the end-effector. By moving the motors, the tendons can be tightened or loosened to control the robot’s configuration. Unlike conventional revolute joints, where each joint is actuated directly by a motor, the tendon-driven architecture leads to a non-unique set of cable tensions that can maintain the robot at a given configuration.

Sensor measurements are available from force sensors on each tendon, and a camera-based positioning system giving the pose of each system link. The motors tightening the tendons can be controlled by setting position or velocity references, or by directly specifying motor currents to control the motor torque.

\vspace{-4pt}
\subsection{Kinematics}
\label{sec:constant_ratio}
\vspace{-4pt}

\begin{figure}[htbp]
\centerline{\includegraphics[width=\linewidth]{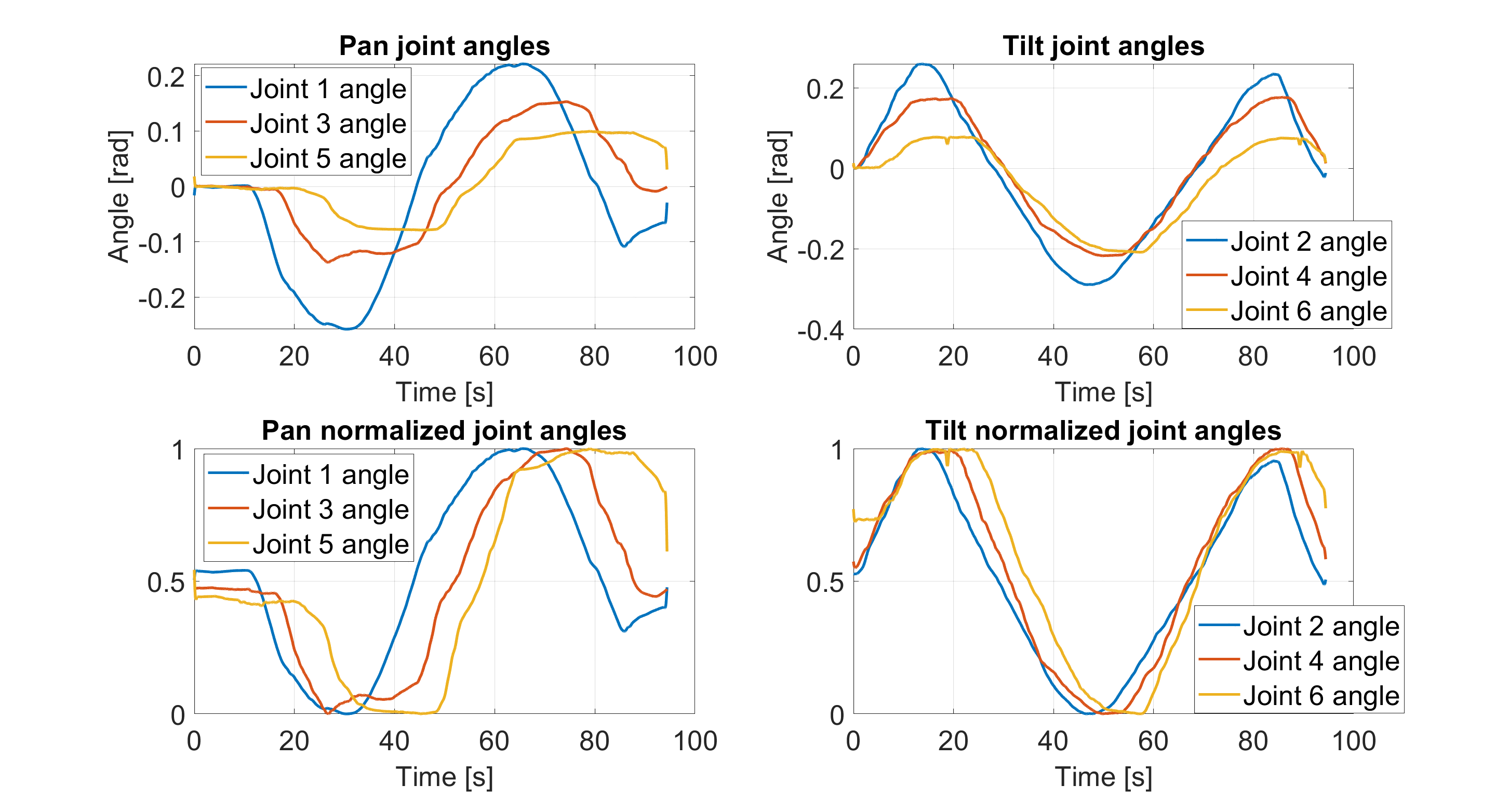}}
\caption{Pan (left) and tilt (right) joint angle displacements (upper) and their normalized values (lower).}
\label{fig:constant_ratio}
\end{figure}

The kinematics of the CERN snake robot were derived in \citep{hansen2025}, using the constant curvature assumption \citep{rao2021} for the inverse kinematics. Experimental data collected using a camera motion tracking system for measuring the joint angles reveal that joints closer to the base exhibit larger angular displacements than the joints farther from the base, shown in Fig.~\ref{fig:constant_ratio}. To compare joints, the data were normalized by dividing each joint angle by its maximum value. The normalized plots reveal that joints in the same plane follow a similar pattern, indicating approximate co-linearity. This motivates a modification of the constant curvature assumption.

Rather than assuming a constant curvature, we assume a constant ratio between joint angles in the same plane, derived from the experimentally observed extremes:
\begin{equation}
\begin{split}
    q_3 &\approx 0.6493 q_1, q_5 \approx 0.2053 q_1,\\
    q_4 &\approx 0.6442 q_2, q_6 \approx 0.2291 q_2,
\end{split}
\end{equation}
where $q_i$ is the angle of joint $i$. This ratio-based approximation effectively captures the robot's configuration using only two variables in most operational scenarios. However, it does not capture all dynamics when there are external forces acting on the middle links, or external torques on any of the links. Still, the assumption that the system state can be accurately described with a single coordinate for each rotational axis aligns with standard modeling approaches for continuum- \citep{rao2021}, and rolling-joint systems \citep{kim2014}.

In addition to reducing model complexity, this kinematic dependency between the joint angles in the same planes also supports the choice of using the first two joint angles as the output of the data-driven dynamic model. Including all joints could introduce co-linearities in the outputs, potentially leading to linear dependencies in the learned system matrices, which may degrade the performance of some identification methods.

 \subsection{Dynamics}
A dynamic model of the system was developed in \cite{dantuono2023}, and reformulated for control-oriented purposes in \cite{dantuono2024}, taking the standard form
\begin{equation}
    \bm{B}(\bm{q})\ddot{\bm{q}} + \bm{n}(\bm{q}, \dot{\bm{q}}) = \bm{\tau}.
\end{equation}
Here, $\bm{B} \in \mathcal{R}^{N \times N}$ is the system inertia matrix, $\bm{n} \in \mathbb{R}^N$ represents the Coriolis/centripetal effects and gravitational forces, and $\bm{\tau} \in \mathbb{R}^N$ is the control input, consisting of the generalized torque at each joint.

However, this model does not accurately capture certain complex dynamics that is difficult to model, such as the friction affecting the tendons while moving through the robot. While the model provides valuable qualitative insight and is useful for simulations, comparisons of experimental and predicted model data reveals a lack of accuracy; the model is too inaccurate to apply directly for model-based control. Moreover, as discussed in the previous section, the joint angles oriented in the same plane are kinematically dependent on each other during normal operation. Calculating the dynamics of each joint independently can therefore introduce redundancy and inefficiency. In the worst case, simulations using this full model may evolve into configurations that are not physically realizable or kinematically consistent.


To address these challenges and improve computational efficiency, we seek to develop a reduced-order model that better reflects the robot's true behavior using data-driven system identification methods. This approach allows us to capture the essential system dynamics without explicitly modeling all physical interactions, while still maintaining sufficient accuracy for control and prediction.

\section{Data-driven modeling}
\label{sec:datadriven}
This section gives a brief introduction to the modeling approaches compared in the work, namely N4SID, SINDYc and ARX. The three considered methods all attempt to find a discrete-time representation of system dynamics, but differ significantly in how this is achieved and in which form the model is. 

\subsection{Numerical algorithms for subspace state space system identification (N4SID)}
\label{sec:n4sid}
The Numerical Subspace State Space System Identification (N4SID) algorithm \citep{vanoverschee1994} is a robust method for deriving the state-space representation of a discrete-time system of the form
\begin{equation}
\label{eq:discrete_system}
\begin{split}
    \bm{x}_{k+1} &= \bm{A} \bm{x}_k + \bm{B}\bm{u}_k \\
    \bm{y}_{k} &= \bm{C} \bm{x}_k + \bm{D}\bm{u}_k,
\end{split}
\end{equation}
from input-output data. Here $\bm{x} \in \mathbb{R}^n$, $\bm{u} \in \mathbb{R}^p$, and $\bm{y} \in \mathbb{R}^q$. 

The algorithm projects future outputs onto the row space of past data, yielding a matrix that approximates the product of the extended observability matrix and the state sequence. A singular value decomposition (SVD) of this matrix provides a low-rank factorization from which both the system order, observability matrix, and the state sequence 
\(
\bm{X} = \begin{bmatrix}
    \bm{x}_1 & \bm{x}_2 & \cdots & \bm{x}_{m-1}
\end{bmatrix}
\)
are obtained. Notably, N4SID estimates $n$ internal states without requiring direct state measurements, in contrast to the following methods where $y = x$ and thus $n = q$ is required.

By constructing the matrix $\bm{X}'$, which is the state trajectory shifted by one time step, the system matrices $\bm{A}, \bm{B}, \bm{C}$, and $\bm{D}$ can be found by solving the set of linear equations
\begin{equation}
    \begin{bmatrix}
        \bm{X}' \\ \bm{Y}
    \end{bmatrix} =
    \begin{bmatrix}
        \bm{A} & \bm{B} \\
        \bm{C} & \bm{D}
    \end{bmatrix}
    \begin{bmatrix}
        \bm{X} \\ \bm{\Upsilon}
    \end{bmatrix}
\end{equation}
for $\bm{A}, \bm{B}, \bm{C},$ and $\bm{D}$. Here,
\begin{equation}
    \label{eq:upsilon}
    \bm{\Upsilon} = \begin{bmatrix}
        \bm{u}_1 & \bm{u}_2 & ... & \bm{u}_{m-1}
    \end{bmatrix},
    \quad
        \bm{Y} = \begin{bmatrix}
        \bm{y}_1 & \bm{y}_2 & ... & \bm{y}_{m-1}
    \end{bmatrix}
\end{equation}
are matrices consisting of the control inputs, system measurements at each time step, respectively. 


\subsection{Sparse Identification of Nonlinear Dynamics with Control (SINDYc)}
\label{sec:SINDY}
SINDY \citep{brunton2016} is a framework for learning sparse nonlinear dynamics from measured data. Its extension, SINDYc \citep{brunton2016a}, allows learning the dynamics of systems with external inputs. Given sets of $m$ state measurements and control inputs $\bm{X}, \bm{X}',$ and $\bm{\Upsilon}$, SINDYc attempts to learn a nonlinear relation of the form 
\begin{equation}
    \bm{X}' = \bm{\Theta}(\bm{X}, \bm{\Upsilon}) \bm{\Xi}.
\end{equation}
Here $\bm{\Theta}(\bm{X}, \bm{\Upsilon})$ is a library of candidate nonlinear functions, for example
\begin{equation}
    \bm{\Theta}(\bm{X}, \bm{\Upsilon}) =
    \begin{bmatrix}
        \bm{X} & \bm{X}\bm{X}^T & \bm{X}\bm{\Upsilon}^T... &\sin(\bm{X}) ....
    \end{bmatrix},
\end{equation}
where $\sin(\bm{X})$ denotes the elementwise sine of $\bm{X}$, and $\bm{\Xi}$ is a sparse matrix of coefficients. These coefficients are found by solving the optimization problem
\begin{equation}
    \label{eq:sindy_optimization}
    \bm{\xi}_k = \text{argmin}_{\bm{\xi}_k} \norm{\bm{X}'_k - \bm{\xi}_k \bm{\Theta}^T(\bm{X}, \bm{\Upsilon}) }_2 + \lambda \norm{\bm{\xi}_k}_1,
\end{equation}
for each row $\bm{\xi}_k$ of $\bm{\Xi}$, using a sequential thresholded least-squares procedure \citep{brunton2016}. Here $\lambda$ is a hyperparameter chosen to promote sparsity of $\bm{\Xi}$, and $\bm{X}'_k$ is the $k$-th row of $\bm{X}'$.


As with similar approaches, such as EDMD, finding a suitable set of nonlinear functions is crucial. If a nonlinear relationship exists and it is not represented in the function library $\bm{\Theta}$, it will not appear in the learned model. The sparsity-promoting optimization procedure ensures that terms with little or no effect are excluded from the final model, improving the model efficiency. 


\subsection{Automated Regression with eXogenous inputs (ARX)}
\label{sec:arx}
An ARX model belongs to a family of system models known as parametric models \citep{ljung1999}. These models are of the form
\begin{equation}
    \label{eq:parametric_model}
    \bm{x}_k = \bm{G}(q, \bm{\theta}) \bm{u}_k + \bm{H}(q, \bm{\theta}) \bm{e}_k,
\end{equation}
where $q$ is the timeshift operator, $\bm{\theta}$ are the model parameters, and $\bm{e}$ is a disturbance signal. Among the parametric models in \cite{ljung1999}, we also invesigated NLARX, autoregressive moving-average (ARMAX), Output-Error, Box-Jenkins and Hammerstein-Wiener models, but they were all found to give worse performance for our system than ARX. Therefore, we chose to focus our attention on this method. For ARX models, \eqref{eq:parametric_model} can be written:
\begin{align}
    \label{eq:arx_equation}
    \bm{x}_k &+ a_1\bm{x}_{k-1} + a_2\bm{x}_{k-2} + .... + a_{n_a} \bm{x}_{k-n_a} \notag\\
    &= b_1\bm{u}_{k-1} + .... + b_{n_b} \bm{u}_{k-n_b} + \bm{e}_k,
\end{align}
with $e_k \in \mathbb{R}^n$. This can be expressed more compactly as
\begin{equation}
\bm{x}_k = \frac{\bm{B}_{\mathrm{arx}}(q)}{\bm{A}_{\mathrm{arx}}(q)} \bm{u}_k + \frac{1}{\bm{A}_{\mathrm{arx}}(q)} \bm{e}_k,
\end{equation}
with \begin{align}
    \bm{A}_{\rm arx}(q) &= 1 + a_1 q^{-1} + a_2 q^{-2}... + a_{n_a} q^{-n_a}  \\
    \bm{B}_{\rm arx}(q) &= b_1 q^{-1} + b_2 q^{-2} + ... + b_{n_b} q^{-n_b}.
\end{align}
The terms $n_a, n_b$ are design parameters that govern the dimensions of the identified model. 
The ARX model consists of an autoregressive term, which linearly depends on its own previous values, and an exogenous term representing the control input, the model parameters are identified using linear regression.




\section{Experiments}
\label{sec:experiments}
In this section we present the experimental setup used to gather system data, how the models were implemented and validated against real robot measurements, and how the models were used to reconstruct the entire system state. Finally, we implement nonlinear and linear MPCs using the models to assess the real-time control performance on two selected models. 

\subsection{Data collection}
For the implementation of the system identification algorithms, data were collected from the robot prototype at CERN. The prototype was controlled using an open-loop kinematic controller based on \citep{hansen2025} to disambiguate the effects of a feedback controller. 
The system performed a series of circular end-effector trajectories, starting with a radius of 1 cm, and increasing by 1 cm increments up to a circle with a 12 cm radius. Additionally, the robot was driven pseudo-randomly throughout the workspace to capture high-frequent dynamics and to avoid overfitting to the circular motion.

The experiments were conducted using a one-section, six-link prototype of the robot (see Fig.~\ref{fig:snake}). The tendon displacements were actuated by four Dynamixel XH430-W210-R motors, and the corresponding cable forces were measured using Flintec Y1 force sensors. Motion capture was done with Vicon Tracker 3.10.2 with four Vicon Vero grayscale cameras to measure the angle displacements for each joint. The cameras tracked reflective markers that were attached to each link. The cameras captured the entire workspace of the robot with an error margin of approximately $0.3$ mm. Due to some measurement noise, the joint angle measurements were low-pass filtered with a cut-off frequency of $0.001$ rad per sample, with a sampling time of $30$ ms.

The input to the system identification algorithms consisted of the four cable force measurements, assuming that each cable force can be controlled independently. The output was the first two joint angles, obtained from the motion capture system.  Implementation details for the different methods are given in Appendix~\ref{app:implementation}.


\subsection{Model validation}
To validate the identified models, their performance was evaluated on a dataset that was not used for identification. The validation dataset consisted of similar motions to the training set, including circular end-effector trajectories and pseudo-random movements throughout the workspace. The results of the validation are shown in Tab.~\ref{tab:sysID_accuracy}, using the normalized fit percentage as the performance metric, defined as $\text{Fit} = 100 \left(1 - \frac{\norm{y - \hat{y}}}{\norm{y - \bar{y}}} \right),$
where $y$ are the state measurements, $\hat{y}$ the predicted measurements, and $\bar{y}$ the mean of the measured data.
For multi-output systems, the reported fit value represents the mean fit computed across all output channels.

\begin{table}[h]
    \centering
    \renewcommand{\arraystretch}{1.2} 
    \setlength{\tabcolsep}{8pt} 
    \begin{tabular}{lccc}
        & \textbf{SINDYc} & \textbf{ARX} & \textbf{N4SID} \\
        \midrule
        \textbf{Fit (\%)} & 67.74 & 44.90 & 39.12 \\
        \bottomrule
    \end{tabular}
    \vspace{2pt}
    \caption{Accuracy comparison of different system identification methods.}
    \label{tab:sysID_accuracy}
\end{table}

As shown in Table I, the SINDYc method achieved the best performance, with an NRMSE of 67.74\%. While the linear methods ARX and N4SID also showed moderate accuracy, SINDYc outperformed them, benefiting from its ability to capture nonlinear system dynamics while maintaining a sparse model structure.

Attempts to improve the ARX model by introducing nonlinear terms did not yield better performance, suggesting that any nonlinearities present cannot be easily captured in an autoregressive structure.



\subsection{State reconstruction}
As the learned models only include the first two joint angles, it is also necessary to investigate the accuracy of the constant joint ratio approximation used to estimate the remaining joint angles and thus to get a full system description. Based on the results in Tab.~\ref{tab:sysID_accuracy}, the SINDYc model showed the best overall performance and was therefore selected as the basis for estimating the full joint configuration.

\begin{figure}[htb!]
    \centering
    \includegraphics[width=\linewidth,height=6.5cm]{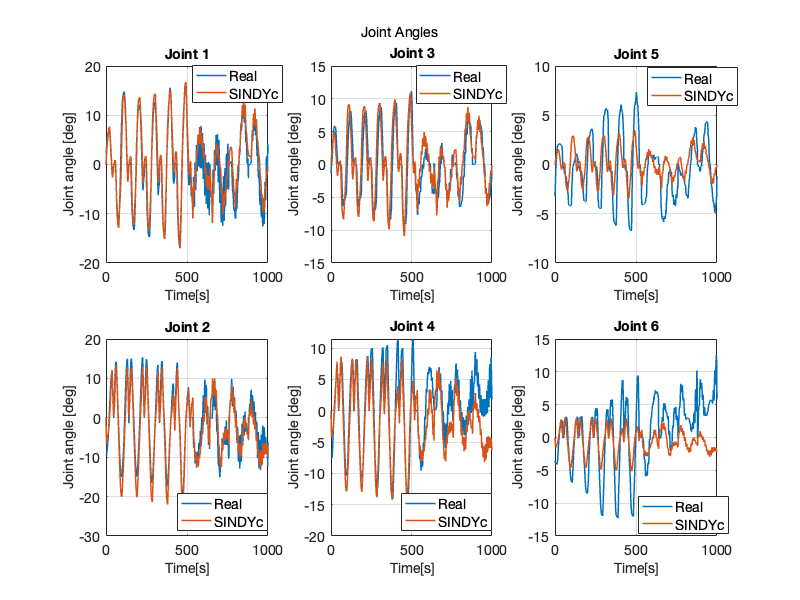}
    \vspace{-8mm}
    \caption{Simulated joint angles using the 2 DoF SINDYc model. Planar joints with odd index, pitch joints even.}
    \label{fig:joints_sindy}
    \vspace{-2mm}
\end{figure}

Fig.~\ref{fig:joints_sindy} shows the performance of the SINDYc algorithm on the validation dataset. The two first joint angles, which serve as outputs of the learned model, are shown in the left column. The four subsequent joint angles are reconstructed using the constant-ratio assumption discussed in Sec.~\ref{sec:constant_ratio} and are shown in the center and right columns.

\begin{figure}[htb!]
    \centering
    \includegraphics[width=\linewidth,height=6cm]{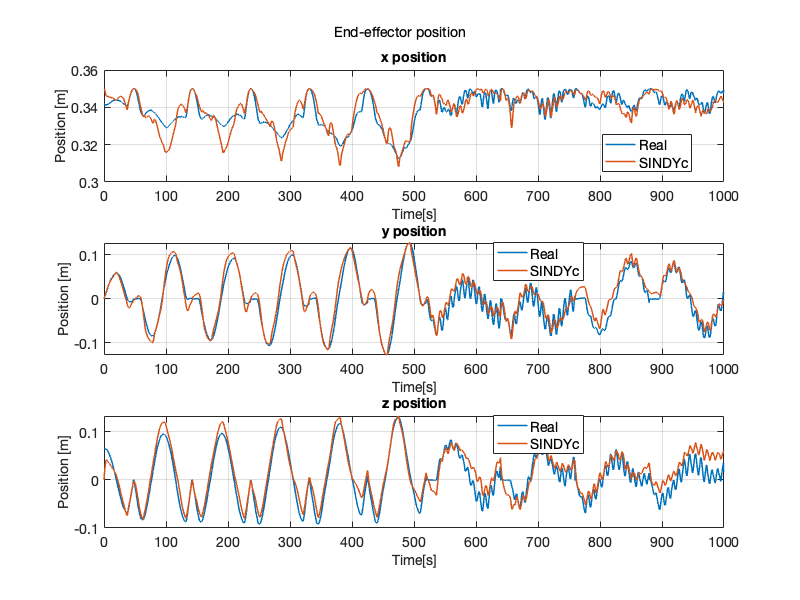}
    \vspace{-2em}
    \caption{End effector comparison of SINDYc simulation vs validation data}
    \label{fig:position_comparison}
\end{figure}

The simulated joint angle values are further converted to Cartesian end-effector positions using the forward kinematics in \citep{hansen2025}. The comparison between the simulated end-effector trajectory and the ground truth from the validation data is shown in Fig.~\ref{fig:position_comparison}. The average Euclidean error between the simulated and measured end-effector positions was $2.48$ cm.

\subsection{Real-time control}
\label{sec:mpc}
To further validate the learned models and assess their feasibility for real-time control, experiments were conducted using the models within a model predictive control (MPC) framework. The nonlinear SINDYc model was tested in the controller, and for comparison, a simpler and more computationally efficient linear MPC was also implemented using the identified N4SID model. Both controllers were implemented using the MPC toolbox in MATLAB. 

\begin{figure}
\centerline{\includegraphics[width=0.95\linewidth]{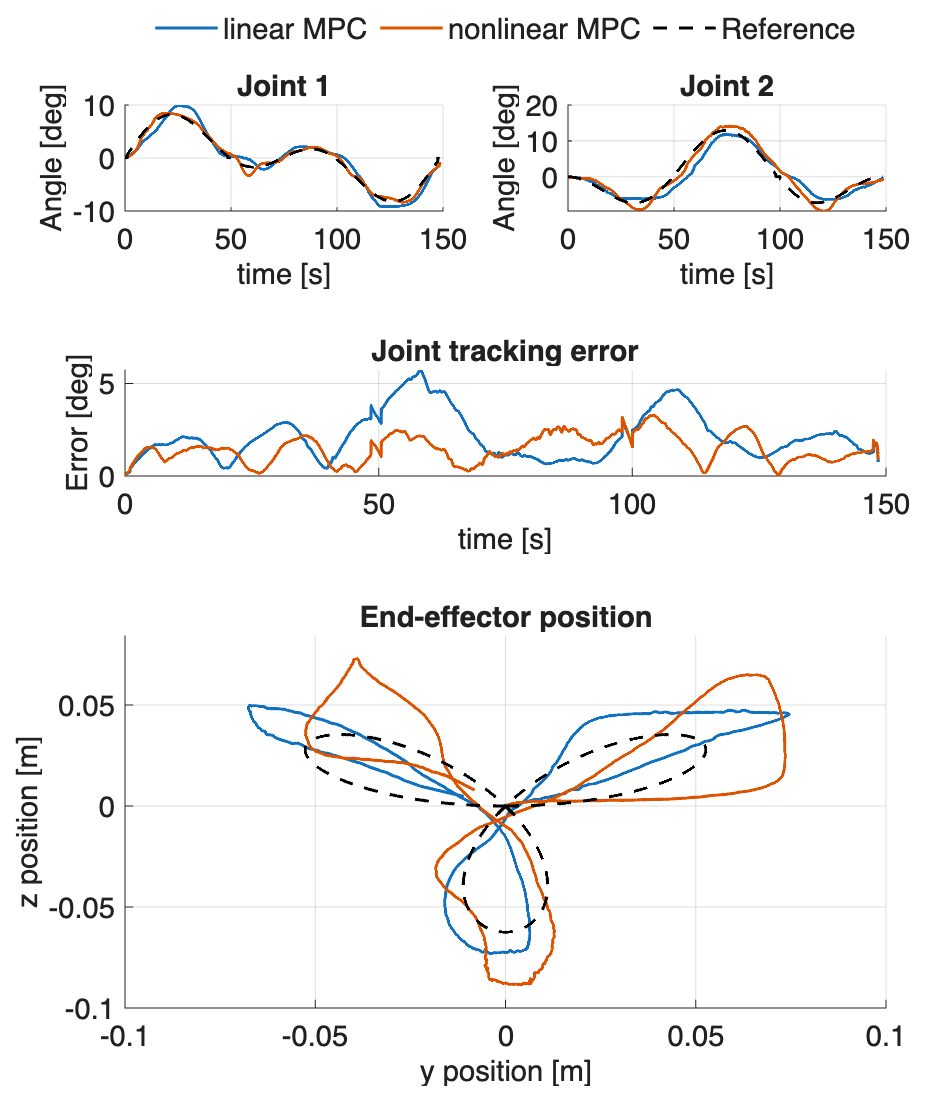}}
\caption{End-effector tracking performance using MPC.} 
\label{fig:mpc_petal}
\end{figure}

The control task was trajectory tracking of the first two joint angles. To evaluate generalization to other motions than the ones used for training the model, a reference trajectory for the end-effector was chosen as a petal-shaped curve, shown in Fig.~\ref{fig:mpc_petal}. The corresponding joint angle references were obtained by converting the end-effector reference position using the constant ratio assumption in Sec.~\ref{sec:constant_ratio}. The cost function for the MPC was defined as
\begin{equation}
    J = \sum_{k=0}^{N-1} \left( \left\|\tilde{\bm{x}}_k^T\right\|_{\bm{Q}}^2 + \left\|\dot{\bm{u}}_k^T\right\|_{\bm{R}}^2  \right) + \left\|\tilde{\bm{x}}_N^T\right\|_{\bm{Q}_f}^2,
\end{equation}
where $\bm{\tilde{x}}_k$ denotes the deviation from the reference trajectory, and $\dot{\bm{u}}$, the rate of change of the control input. Since keeping the motors stationary requires no extra effort regardless of the cable tension, no penalty was applied to the absolute value of the control input $\bm{u}$, only its rate of change $\dot{\bm{u}}$, approximated to $\bm{u}_k - \bm{u}_{k-1}$. The MPC weights and constraints were chosen as $\bm{R} = 0.1 \bm{I}_{4}$, $\bm{Q} = \bm{Q}_f = \bm{I}_2$, input bounds $20~\mathrm{N} < u_i < 190~\mathrm{N}$, and state bounds $-1~\mathrm{rad} < x_i < 1~\mathrm{rad}$, for each element of $\bm{u}$ and $\bm{x}$. The controller frequency was set to $33$ Hz, limited by the force sensor transmission rate. The nonlinear MPC differed from the linear MPC only in the model prediction, leaving the cost function identical for the two cases.


The joint tracking performance and end-effector position is shown in Fig.~\ref{fig:mpc_petal}. Both the linear and nonlinear MPC manage to track the desired joint trajectories well, with slightly better performance from the more accurate, SINDYc model. Much of the error can be attributed to the low-level tendon force controller which was unable to completely track the desired tendon forces, due to both slow motor dynamics and coupling effects between tendons.

It should be noted that the implemented controller does not use feedback from the end-effector position, only from the joint angles. Thus, any kinematic errors from the constant ratio assumption are not corrected during control. The accuracy of the learned methods should therefore mainly be judged on the accuracy of tracking the first two joints, shown in the top of Fig.~\ref{fig:mpc_petal}, not the end-effector position. While including end-effector feedback in the controller would improve performance, as our goal is to validate the learned dynamic models, we chose to keep the controller as simple as possible, and will instead further examine the system kinematics in future work.


The end-effector position does not show the same level of performance as the joint tracking, implying that the constant ratio assumption is not completely accurate for more complex system motions. Especially the two outermost joints show some discrepancy, as seen in Fig.~\ref{fig:joints_sindy}. Nevertheless, the end-effector successfully captures the overall petal shape despite the intricate trajectory. For tasks requiring more accurate end-effector positioning, the controller could be augmented to include feedback of the end-effector position, and thus compensate for the kinematic uncertainties.

\section{Conclusion and future work}
\label{sec:conclusions}
This paper presented the development and first experimental validation of a data-driven 2-DoF dynamic model for a tendon-actuated manipulator with rolling joints developed at CERN. Several system identification techniques—both linear and nonlinear—were implemented and compared. Among these, the SINDYc algorithm achieved the best performance, yielding a joint angle accuracy of 67.74\% and an average Euclidean end-effector error of 2.48 cm. The remaining joint angles were estimated using a constant-ratio assumption based on experimental data, enabling full-body reconstruction from a reduced-order model. The identified models were successfully applied in the design of a model predictive controller (MPC), demonstrating their suitability for real-time, model-based control. 


Future work includes extending the methodology to multi-section continuum robots, incorporating adaptive system identification for dynamic environments, and exploring the integration of learned models into feedback control strategies beyond MPC. Alternative modeling approaches such as deep neural networks could also be investigated. Future developments on the control side also require updating the low-level tendon-force controller, which seems to be the limiting factor for control performance at the moment. Further refinement of the joint angle estimation technique may also enhance reconstruction accuracy in scenarios with external disturbances.

\vspace{-4pt}
\appendix
\section{Implementation details}    
\label{app:implementation}

SINDYc was implemented based on \citep{brunton2017}. After a coarse search through powers of ten, followed by a more refined parameter sweep, $\lambda=0.0035$ was found to provide a good balance between sparsity and model accuracy, resulting in the model presented below. The  MATLAB System Identification Toolbox was used for implementing N4SID, with output $y\in R^2$ being the first two joint angles. A model order of 8 was found during identification. 

The ARX model was implemented using the same toolbox, with parameters set as $n_a = 8 * {\rm ones}(n_y, n_u), n_b = 8 * {\rm ones}(n_u, n_y),$ and $ n_k = 1 * {\rm ones}(n_y, n_u) $. These parameters were found using the model order of 8 from N4SID as a baseline, and through trial and error seemed to give the best performance.


\vspace{-6pt}

\bibliography{Datadriven}
\end{document}